\documentclass{article}

\usepackage{natbib}
\usepackage[preprint]{neurips_2019}

\usepackage[utf8]{inputenc} 
\usepackage[T1]{fontenc}    
\usepackage{hyperref}       
\usepackage{url}            
\usepackage{booktabs}       
\usepackage{amsfonts}       
\usepackage{nicefrac}       
\usepackage{microtype}      
\usepackage{mathtools,etoolbox}
\DeclarePairedDelimiterX{\abs}[1]{\lvert}{\rvert}{\ifblank{#1}{{}\cdot{}}{#1}}

\usepackage{caption}
\usepackage{subcaption}
\usepackage{booktabs, multirow} 
\usepackage{soul}
\usepackage[table]{xcolor} 
\usepackage{changepage,threeparttable} 
\usepackage{filecontents,catchfile} 
\usepackage{numprint}
\npdecimalsign{.}
\nprounddigits{2}

\usepackage{amsmath}
\usepackage{bm}
\usepackage{pdfpages}
\usepackage{graphicx}

\author{%
  Niall Taylor\textsuperscript{\textnormal{1}}\footnotemark[1]\qquad
  Yi Zhang\textsuperscript{\textnormal{1}}\thanks{These authors contributed equally to this work.}\footnotemark[1]\qquad 
  Dan W Joyce\textsuperscript{\textnormal{1,2}}\qquad \\
  \textbf{Alejo Nevado-Holgado}\textsuperscript{\textnormal{1}}\qquad
  \textbf{Andrey Kormilitzin}\textsuperscript{\textnormal{1}}\qquad\\ \\
  \textsuperscript{1}Department of Psychiatry, University of Oxford, Oxford, OX3 7JX, UK \\
  \textsuperscript{2}NIHR Oxford Health Biomedical Research Centre, Oxford, OX3 7JX, UK \\ \\
  \texttt{\{first\_name\}.\{last\_name\}@psych.ox.ac.uk} \\
}

\title{Clinical Prompt Learning with \\Frozen Language Models}

\begin{document}

\maketitle

\begin{abstract}
     Prompt learning is a new paradigm in the Natural Language Processing (NLP) field which has shown impressive performance on a number of natural language tasks with common benchmarking text datasets in  full, few-shot, and zero-shot train-evaluation setups. Recently, it has even been observed that large but frozen pre-trained language models (PLMs) with prompt learning outperform smaller but fine-tuned models. However, as with many recent NLP trends, the performance of even the largest PLMs such as GPT-3 do not perform well on specialized domains (e.g. medical text), and the common practice to achieve State of the Art (SoTA) results  still consists of pre-training and fine-tuning the PLMs on downstream tasks. The reliance on fine-tuning large PLMs is problematic in clinical settings where data is often held in non-GPU environments, and more resource efficient methods of training specialized domain models is crucial.    
     We investigated the viability of prompt learning on clinically meaningful decision tasks and directly compared with more traditional fine-tuning methods. Results are partially in line with the prompt learning literature, with prompt learning able to match or improve on traditional fine-tuning with substantially fewer trainable parameters and requiring less training data. We argue that prompt learning therefore provides lower computational resource costs applicable to clinical settings, that can serve as an alternative to fine-tuning ever increasing in size PLMs. Complementary code to reproduce experiments presented in this work can be found at: \url{https://github.com/NtaylorOX/Public_Clinical_Prompt}

\end{abstract}


\section{Introduction}
\label{sec:intro}

The field of Natural Language Processing (NLP) has seen a surge in the use of deep learning in recent years, partly due to the increased capacity and availability of powerful GPUs and cloud computing globally. Both academic and industry research have subsequently become dominated by the use of large Pretrained Language Models (PLMs), which are typically commercially produced and trained on enormous amounts of text data in a self-supervised manner through language modelling objectives such as Masked Language Modeling (MLM) and next word prediction. Two major PLM families are the bidirectional encoder representations from
transformers (BERT)  \cite{devlin-etal-2019-bert} which originally had ~110 million trainable parameters, and Generative Pre-trained Transformer 3 (GPT-3)  \cite{radford2019language, brown2020language} and the new Meta's Open Pre-trained Transformer Language Model (OPT)   \cite{zhang2022opt}, with $\sim$ 175 billion parameters. With these PLMs one can fine-tune on new domains and design downstream tasks with relative ease, often resulting in state of the art results on a number of popular datasets and tasks   \cite{devlin-etal-2019-bert, Lester2021-power-scale-prompt}. However, "out of the box" PLMs typically do not perform well on out-of-domain texts   \cite{robust-transfer-learning}: Thus taking a BERT model trained on non-medical texts and applying it to a niche medical text domain often leads to a lackluster performance  \cite{BioBERT, huang2019clinicalbert}. Instead domain specific PLMs are often created through continued pre-training on domain specific corpora when available   \cite{alsentzer-etal-2019-publicly, peng2019transfer, gururangan-etal-2020-dont, senior2020identifying, vaci2021real}. Moreover, to then leverage the knowledge of these domain specific PLMs to achieve a downstream task requires further training of a task-specific module, such as a classification head, attached to the end of the PLM   \cite{devlin-etal-2019-bert, wolf-etal-2020-transformers}. Typically downstream task fine-tuning requires further training of all of the PLMs parameters, in addition to the attached task specific head(s).

 This fine-tuning approach is suitable when the application domain has an abundance of text data, which in many situations is not feasible. For instance in clinical settings, there are major data privacy issues and consequently large open medical datasets are difficult to produce. On top of this, the written language used in clinical text can differ drastically to that of the same language found in general written texts, and even between clinical institutions  \cite{huang2019clinicalbert, 2015-clinical-nlp-challenging, kormilitzin2021med7}. Together this makes creating general purpose clinical PLMs quite difficult. Additionally, the NLP community has seen a trend of increasing model size to enhance performance; Microsoft recently produced a monolithic 530 billion parameter model named Megatron for state of the art performance on generative tasks   \cite{megatron-turing-2022}. Whilst impressive, to utilise such models for specific domains of interest will likely require full or partial fine-tuning, which has the massive computational, financial investment and of course, environmental impacts  \cite{bender2021dangers}. 

Regardless of the size issues of the PLMs, there is still a real benefit in their application to new domains and downstream tasks through traditional fine-tuning, including the biomedical domain   \cite{huang2019clinicalbert, alsentzer-etal-2019-publicly,van-aken-etal-2021-clinical}. The persistent concern is the need to fine-tune both the entire PLM and task specific head to produce viable performance on many tasks. In the case of the recently produced super large PLMs, this can require the continual training of models that require large suites of high end GPUs, with proportional financial costs. GPUs and high-performance computing clusters are rarely available to hospitals and community clinics that hold much of the existing medical data. Further to this, traditional fine-tuning can lead to a very specific fine-tuned model that is now very far from its initial pre-trained state, which may cause catastrophic forgetting of the pretrained knowledge  \cite{chen-etal-2020-recall}. Fine-tuning has also been reported to exploit spurious correlations of the smaller domain-specific dataset, damaging its generalizability  \cite{gururangan-etal-2018-annotation, niven-kao-2019-probing}. We have also observed this lack of generalizability in medical text when fine-tuning and then validating across American and British English  \cite{hofer2018few}. Considering the limitations outlined above, we recognise there is now a movement in the NLP community back towards resource efficient training regimes and models to avoid the need for full scale domain specific training. One promising strategy is known as prompt learning, which aims to close the design gap between the PLMs training objectives and downstream tasks by reformulating the downstream tasks as language modelling objectives   \cite{prefix-tuning,pre-train-prompt-predict}. Prompt learning has evolved from earlier works which have reformulated all NLP downstream tasks as text-to-text tasks   \cite{t5-paper} and more recently using task examples within the input text as a form of prompt in auto-regressive PLMs   \cite{brown-lms-few-shot}. An exciting direction in the prompt learning research space has been its potential in few-shot or low resource settings, relying on frozen PLMs  \cite{tsimpoukelli2021multimodal} instead of fine-tuning them: The number of parameters to train decreases dramatically when using frozen PLMs and thus reduces computational requirements  \cite{lu2021pretrained}. The major gap in the literature is in the application of prompt learning to clinical or biomedical datasets, and in particular clinical support tasks.

We explore the suitability and performance of prompt learning applied to clinical classification tasks with a direct comparison to traditional fine-tuning methods in full and few-shot training scenarios. Our primary focus is on the performance of these approaches when using a frozen PLM, which is desirable for many reasons, but primarily the consequent reduction in training cost and computational resources required to adapt to new domains or downstream tasks. Conceptually we are not introducing a new methodology, rather exploring different applications of prompt learning to the biomedical domain and importantly to clinical tasks, rather than simple natural language probing tasks. We observed that prompt learning strategies can outperform traditional fine-tuning on different clinical tasks in both few-shot and full training scenarios with frozen PLMs. This work can serve as a prompt learning framework for clinical tasks and as a basis for further work in this space.

\section{Related Work}
\label{sec:related}

Since the summer of 2021 there has been a steady influx of research papers concerning prompt learning for common benchmarking open-NLP datasets such as Stanford Sentiment Treebank v2 (SST2), and the General Language Understanding Evaluation (GLUE)   \cite{pre-train-prompt-predict, brown-lms-few-shot, Sanh2021-t0-paper, Lester2021-power-scale-prompt, p2-prompt-tuning-few-shot, prefix-tuning}. The datasets and tasks are standard in the field of NLP, and revolve around natural language understanding (NLU) tasks. The common finding is that prompt learning can reach the performance of traditional fine-tuning, and often outperform in few-shot settings. Although the ability of prompt learning to match performance of traditional fine-tuning seems to scale with PLM size   \cite{p2-prompt-tuning-few-shot}.
One notable paper has investigated the use of GPT-3 for biomedical text datasets in a few-shot setting, finding a decrease in performance when compared to similar tasks in the standard NLU datasets   \cite{gpt-3-poor-few-shot-biomed}. This suggests that even the largest PLMs cannot be applied directly to specialised domains and expect good performance, and that domain specific PLMs are still sought for optimal results. 


Recently, prompt learning was used to investigate the zero-shot performance on a clinical task using different PLMs and manual prompt templates    \cite{health-prompt-paper}. They found that biomedically trained PLMs outpeformed general PLMs for one task, and we hope to extend these findings by introducing different prompt learning training strategies and clinical tasks.

\section{Methods}
\label{sec:methods}

\subsection{Traditional fine-tuning}

Conventional fine-tuning can be achieved by adding task-specific layer(s) or entire multi-layer perceptron (MLPs). The exact approach to processing the PLM output is dependent on the task.  \begin{figure}[ht]
\centering
\includegraphics[scale= 0.75]{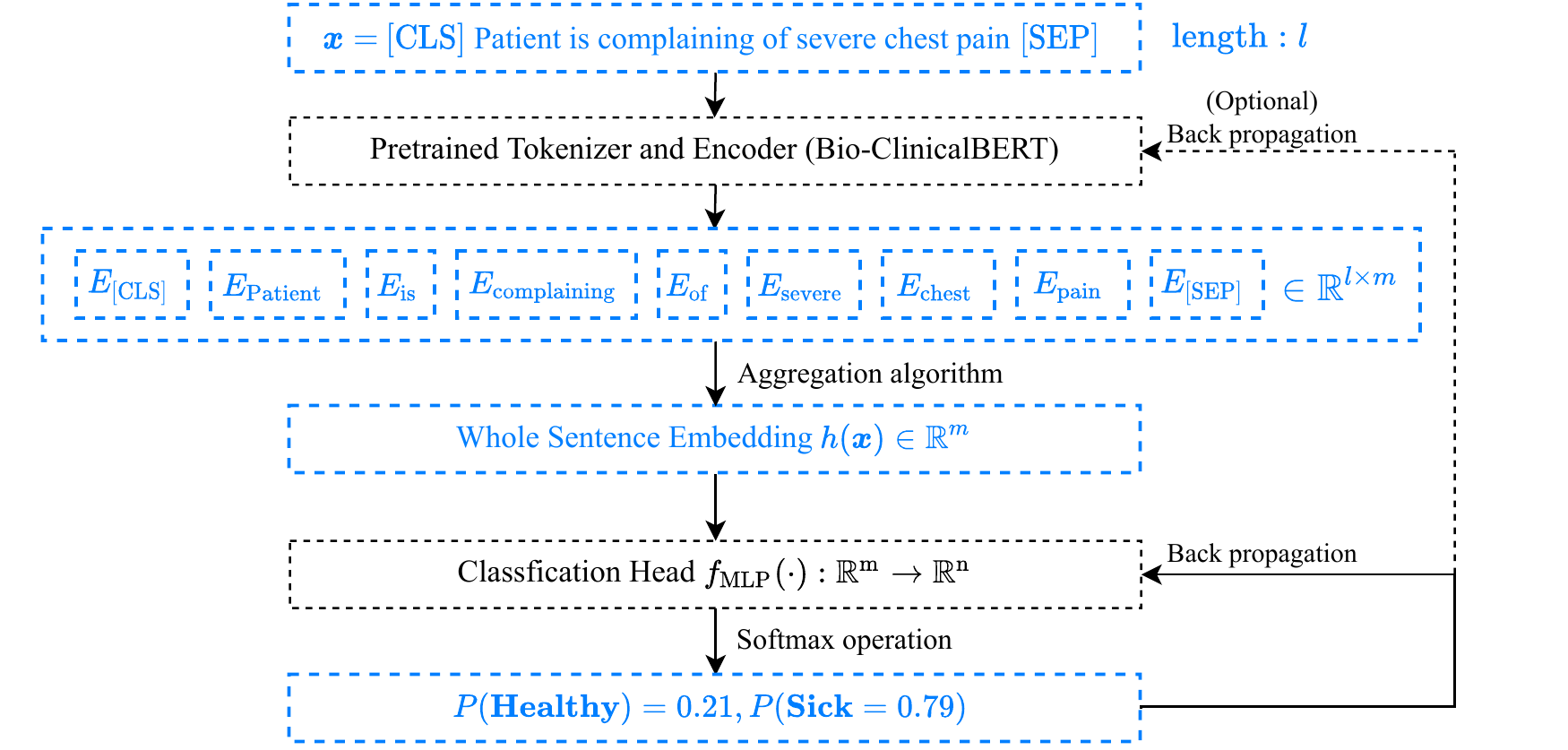}
\caption{Illustration of conventional fine-tuning method, with an option to freeze the PLM, shown in dotted line. Here [CLS] and [SEP] tokens are special tokens for BERT-based models that are added to the beginning and end of sequences.}
\label{fig:fine-tuning}
\end{figure}

In the case of document classification, the downstream task head is an MLP $f_\text{MLP}(\cdot)$ which takes the pooled sentence embedding output by the PLM  as input and generates an $n\text{-dimensional}$ vector, where $n$ is the number of classes. That is, given an input text $\boldsymbol{x}$, we first process the raw input with the PLM to get the $m$- dimensional embedding of each token. Then a pooling operation, such as the mean, as applied to all token embeddings to produce a singular sentence embedding $h(\boldsymbol{x})$ of the same dimension $m$. Then $h(\boldsymbol{x})$ is fed to the MLP block in a standard feed-forward manner to get the probability  across $n$ classes with a softmax operator:
$$ P(y \mid \boldsymbol{x}) = \frac{\exp{ ((f_\text{MLP}\left(h(\boldsymbol{x})\right)_y)}}{\exp{\left( \sum_{i=1}^{n} f_\text{MLP}(h(\boldsymbol{x}))_i \right)}}.$$ The MLP block can have any depth of layers $m \in \mathbb{N}$, while in our experiments, we opted for $d = 2$. Since the additional MLP block and PLMs are modular, their respective parameters are stored separately and we can opt to freeze the parameters of one or the other. An example of processing a short input text sequence using this method is shown in Fig. \ref{fig:fine-tuning}.

\subsection{Prompt Learning}

Generally, prompt learning can be achieved via the following steps: Given an input text $\boldsymbol{x}$, we modify it to a prompt format $\boldsymbol{x'} = f_p(\boldsymbol{x})$, where $f_p$, often called a template, will normally prepend, append, or insert a number of additional token embeddings to the original input along with a masked token, denoted by <[MASK]>. We then feed $\boldsymbol{x'}$ into the PLM to predict the masked token, which is the same as the Masked Language Modelling (MLM) pre-training objective of most BERT-based models. The result of the model will be a distribution over the fixed vocabulary $\mathcal{V}$ of the tokenizer. A second and crucial step is to map tokens or words in the known vocabulary of the PLM to class labels in the downstream task, achieved with a mapping $g: \mathcal{V} \mapsto \mathcal{C}$, where $\mathcal{C}$ is the set of classes. This is known as answer engineering, or verbalization (we will use the term verbalizer and verbalization throughout). The verbalizer can be seen as a mapping between single, or multiple different tokens to distinct class labels. The embedding or hidden state represented at the <[MASK]> position output by the PLM is then passed through a standard language model head, or classifier, and probabilities of the verbalizer derived class label tokens are derived.

\begin{figure}[ht]
\centering
\includegraphics[scale = 0.9]{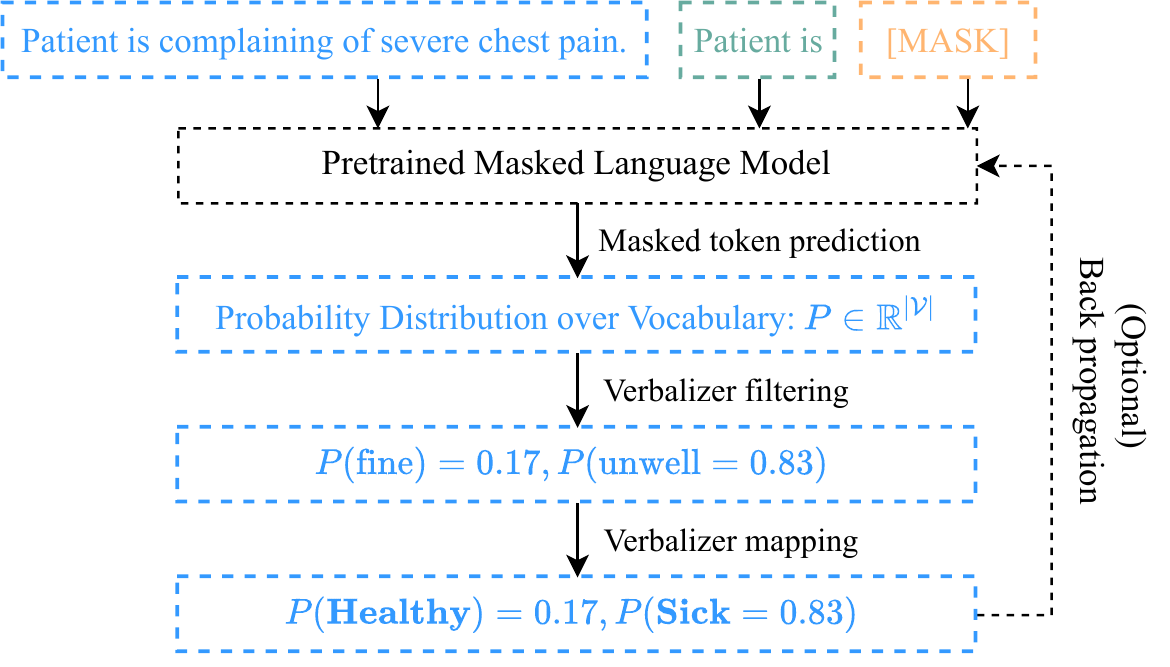}
\caption{Illustration of manual template and verbalizer in prompt learning.}
\label{fig:manual}
\end{figure}

A simple prompt-based clinical classification example could be to determine whether a patient has heart disease with class labels as sick and healthy, a prompt learning setup could be as follows: Take
the template ``<clinical text> <prompt=``Patient is''> <[MASK]>'', where
 <clinical text> represents the original input text, the <[MASK]> token is the label or class to predict. The verbalizer will map certain tokens to each class of sick and healthy separately, essentially a dictionary mapping e.g. \{ ``\textbf{Healthy}'': `fine', and ``\textbf{Sick}'': `unwell'\}. Subsequently if the token predicted at  the <[MASK]> position is 'fine' then this will be mapped to the \textbf{Healthy} class. Thus, the sentence ``Patient is complaining of severe chest pain.'' will
first be wrapped by the pre-defined template as ``Patient is complaining of severe chest pain. Patient is <[MASK]>''. The wrapped sentence is
then tokenized and fed into the PLM to predict the
distribution over vocabulary on the <[MASK]> token
position, although we just care about the probabilities of the tokens (`fine' and `unwell') that are mapped to each of the classes that are contained in $\mathcal{V}$, with ``unwell'' hopefully having a higher probability to be predicted by the masked language model predictor and the class ``sick'' ultimately being predicted. We offer an illustration of the basic prompt framework in Fig. \ref{fig:manual}.

Within the broad prompt learning framework there are important decisions to make about the construction of prompt templates and verbalizers. At its infancy templates were manually created, often based on human knowledge of the task domain, with massive variance in performance with even subtle perturbations of the template and verbalizer   \cite{Lester2021-power-scale-prompt,Hu2021}.

To enable a standardised framework for prompt learning a team have developed OpenPrompt to enable reproducible prompt based research by creating a open source and unified code-base   \cite{Ding2021}. We shall first define the templates and verbalizers used in the framework and our experiments. We refer to the classical prompt learning strategy with handcrafted templates and verbalizers as manual templates and manual verbalizers respectively. This strategy was first proposed as the Pattern-Exploiting Training (PET)  \cite{schick-schutze-2021-exploiting}. We denote the set of words in the verbalizer for each class $y \in \mathcal{C}$ to be $\mathcal{V}^y$. The probability of each class given the input $\boldsymbol{x}$ and its prompt $\boldsymbol{x'}$ is thus:
$$
P(y \mid \boldsymbol{x})=\frac{\exp \left(\frac{1}{\left|\mathcal{V}^{y}\right|} \sum_{t \in \mathcal{V}^{y}} P_{M}\left(t \mid \boldsymbol{x'}\right)\right)}{\sum_{i=1}^{|\mathcal{C}|} \exp \left(\frac{1}{\left|\mathcal{V}^{i}\right|} \sum_{t \in \mathcal{V}^{i}} P_{M}\left(t \mid \boldsymbol{x'}\right)\right)}.
$$

Manual templates and verbalizers are discrete and bounded to the PLMs vocabulary, so there are no extra parameters to train, although fine-tuning the PLM is possible.

The engineering of the manual components of prompt learning is not straight forward, with large variations in performance emerging from small changes to the tokens, and typically domain expertise is required. One can however sacrifice the human interpretability of the manual components and create trainable or soft prompt components. Soft prompt learning operates in the same manner as manual approach, but replaces the fixed manual components with trainable embeddings (continuous vectors) of same dimension as the original PLM. The error from the downstream task can then be back-propagated to tune only the embeddings for the template and verbalizer   \cite{Lester2021-power-scale-prompt}. Normally, a manual template has the form of $\boldsymbol{x'} = \{ \left[ P_{0}, P_{1}, \ldots, P_{j}\right], \boldsymbol{x}, \left[ P_{j+1}, P_{j+2}, \ldots, P_{k}\right], \text{[MASK]} \}$, where for $i \in \{0,1,\ldots,k\}$, $P_i$ denotes the token of the template. And since $\boldsymbol{x'}$ is fed to the PLM to get $h(\boldsymbol{x'})$, the prompt tokens $P_i$ are also mapped to the embedding space, where we can assume $h(P_i)$ to be optimized during training and such tokens are denoted as <[soft]> in the template format. A template where all tokens are <[soft]> is called a soft template, while a template with a mixture of manual and <[soft]> tokens is called a mixed template.

\begin{figure}[h]
\centering
\includegraphics[scale= 0.75]{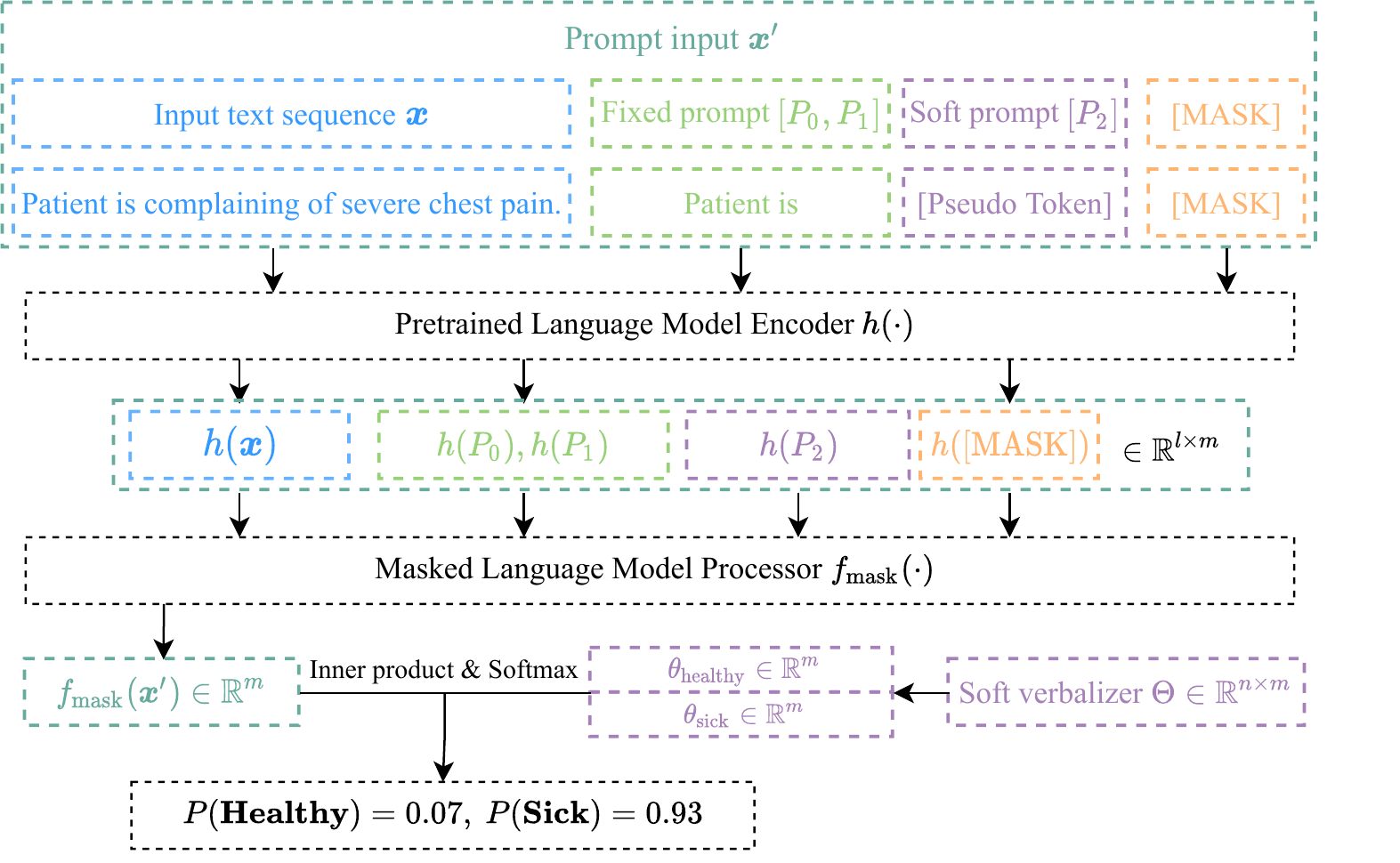}
\caption{Illustration of soft template and verbalizer in prompt learning. If the <[soft]> token $P_2$ is not defined manually in advance, the embedding $h(P_2) \in \mathbb{R}^m$ will be randomly initialized in the hidden space.}
\label{fig:soft}
\end{figure}

Similarly, a soft verbalizer can be assumed as replacing words in verbalizer with trainable vectors for each class. Therefore, when using the soft verbalizer, there is no need to build the mapping from vocabulary $\mathcal{V}$ to class labels $\mathcal{C}$ as the trainable vectors do not have semantic meaning. The resulting verbalizer then becomes a matrix operator $\Theta \in \mathbb{R}^{n \times m}$, where $n$ represents the number of classes and $m$ represents the dimension of generated hidden embeddings. For better understanding, we denote the $i$-th row of $\Theta$ as $\theta_i$ for each trainable vector of class $i$. To compile with the soft verbalizer which takes hidden embeddings from the PLM as input, the original decoder head of the PLM is removed. We denote the resulting mapping from $h(\boldsymbol{x'}) \in \mathbb{R}^{l \times m}$ to the prediction of hidden representation of <[MASK]> as $f_\text{mask}: \mathbb{R}^{l \times m} \rightarrow \mathbb{R}^m$, where $l$ is the sequence length of $\boldsymbol{x'}$. Therefore, the probability of class $y$ given the input $\boldsymbol{x}$ and its prompt $\boldsymbol{x'}$ can be calculated by 
$$
P(y \mid \boldsymbol{x})=\frac{\exp \left( \theta_y^\top f_\text{mask}(h( \boldsymbol{x'}))\right)}{\sum_{i=1}^{n}  \exp \left( \theta_i^\top f_\text{mask}( h(\boldsymbol{x'}))\right)}.
$$

For further details and origins of prompt learning see: P-tuning  \cite{p-tuning}, prefix tuning   \cite{prefix-tuning} and WARP  \cite{hambardzumyan-etal-2021-warp}.

\subsection{Pre-trained Language Model}
 As we wanted to compare the performance of prompt learning and traditional fine-tuning in a best case scenario, we chose the Bio-ClininalBERT   \cite{alsentzer-etal-2019-publicly}. Bio-ClinicalBERT was essentially pre-trained on all MIMIC-III notes and a large collection of PubMed abstracts and full articles by being initialized from weights produced by another biomedical BERT model, BioBERT \cite{BioBERT}. Whilst we appreciate this may be an overly optimized model for the dataset used in this paper, we argue the point of the experiments presented here is to compare and contrast the ability of the different modelling frameworks to leverage what has been learned by a PLM for clinical tasks. As has already been shown extensively, PLMs benefit from domain specific pre-training   \cite{gururangan-etal-2020-dont}, what is lesser known is whether current pre-prompt learning approaches are fully utilising these language models. 

\subsection{Clinical Dataset}
We use the Medical Information Mart for Intensive Care III (MIMIC-III)   \citep{Johnson2016}, an open source medical dataset developed by the MIT Lab for Computational Physiology. It comprises of de-identified health data associated with 38,597 critical care patients and 58,976 intensive care unit (ICU) admissions at the Beth Israel Deaconess Medical Center between 2001 and 2012. Data includes demographics, vital signs, laboratory tests, medications, caregiver notes, imaging reports, and mortality in and out of hospital. The number of possible tasks with this dataset is quite large and varied, but we focus on classification tasks which utilise free text notes alone. Moreover, to allow comparisons with other baselines we derive clinical task datasets used in previous research   \cite{van-aken-etal-2021-clinical,Unsupervised-Pre-Training-on-Patient-Population-graphs,Wang_2020, boag-whats-in-note} as well as deriving our own triage task, described below. An important note is that whilst some of the derived clinical tasks may benefit from utilising the multi-modal data available for each patient, we focus purely on the free text clinical notes.
Full details and code for reproducing these datasets and experiments is provided by authors. \footnote{complementary code to reproduce experiments is provided at: \url{https://github.com/NtaylorOX/Public_Clinical_Prompt}}

\section{Experiments - Clinical tasks}
\label{sec:experiments}

\paragraph{ICD-9 50}
Within the MIMIC-III data and other EHRs are standardised International Classification of Diseases version 9 (ICD-9) codes, which are used to record diagnosis and procedures. A common task is to classify the ICD-9 diagnosis code based on a patients data and automate the whole process, and one can do so from the free text notes alone. There are approximately 2,000 diagnosis codes present in the MIMIC-III dataset, with a very skewed distribution, and a resulting extreme multi-class problem which is beyond the scope of this paper. Thus for our classification task we opt to subset top 50 most frequent ICD-9 diagnosis codes that have a corresponding set of clinical notes, as has been done before \cite{code-synonyms, Wang_2020, van-aken-etal-2021-clinical}. 

\paragraph{ICD-9 Triage task}

A potential concern with the ICD-9 diagnosis code classification is that the codes themselves may be mentioned explicitly in the notes   \cite{van-aken-etal-2021-clinical}\footnote{it was shown samples where diagnosis was not mentioned explicitly only had a slight drop in performance}, and further, simply classifying patients' ICU discharge notes by ICD-9 code lacks ecological validity as a clinical decision support task.  For example, within a hospital setting, patients admitted to an ICU will be treated and then ``stepped down'' (discharged) to another ward or team to progress their treatment when they no longer require ICU. With assistance from clinicians, we therefore designed a novel task that aims to make the classification task more similar to the decision making process of arranging patient flow on discharge from the ICU.  For example, a patient being discharged from the ICU after a cardiac event will likely be ``stepped down'' to a cardiology team.  Similarly, a patient admitted to ICU with obstetric complications will likely be stepped-down to a maternity ward.  In essence we grouped together the ICD-9 diagnosis codes into ``teams'' that reflect the triage or patient-flow decision making found in hospital settings.

For this task we selected the top 20 most frequent ICD-9 diagnosis codes in MIMIC-III and a clinician derived  triage groups based on which team would likely continue the patient's care on being stepped down from ICU.  The training classes are therefore many-to-one mappings of ICD-9 codes to discharge teams and we derived the following seven post-ICU discharge destination teams: Cardiology, Obstetrics, Respiratory Medicine, Neurology, Gastroenterology, Acute or Internal Medicine, and Oncology. The resultant dataset consists of 15,000 clinical notes across the 7 triage categories. 

\paragraph{In hospital mortality}
One of the most frequently used benchmark clinical support tasks with the MIMIC-III dataset is the prediction of whether a patient will survive their hospital episode. Within the MIMIC-III database are structured data relating to the mortality status of a patient, which paired with a date and timestamp allows for easy labelling of the data. Only notes prior to the mortality flag are considered, and some simple regular expression rules were used to filter any notes that had explicit mentions of a patients death, similar to that of previous work   \cite{boag-whats-in-note,van-aken-etal-2021-clinical}.

\paragraph{Length of stay in ICU}
Predicting how long a patient will require ICU is of significant value to hospitals who aim to optimise the flow of patients in resource-limited settings (that is, there are usually very few ICU beds compared to the hospital's overall bed capacity).  We model this as a three way classification task, binning length of stay in the following categories: Under 3 days, 3 to 7 days, 1 week to 2 weeks, more than 2 weeks   \cite{van-aken-etal-2021-clinical}.

\paragraph{Full and few-shot training}
We will be comparing the performance of models in full and few-shot training setups. Validation and test set performance is always carried out on the full validation and test sets to enable direct comparisons in performance. An important note for our few-shot experiments is that sample size will refer to the number of samples per class, i.e. ${N} = {s} \times {c}$ where $N$ is the total training samples, $s$ is the sample size per class and $c$ is the number of unique classes. Note in some instances not all classes can fill the sample size, so for some few-shot experiments there will remain a class imbalance. All results presented are on held-out test sets for each task.

\section{Results}

\subsection{Different prompt learning setups}

The number of possible combinations of templates and verbalizers in the prompt learning framework is vast, and as such we have opted to utilise previous research to derive the most suitable for our use case. To this end we conducted an initial experiment comparing the performance of four prompt learning combinations on one clinical task to establish the best performing combination. We chose the ICD-9 Triage task as the baseline due to it being a relatively straight forward multi-class classification problem and with a reasonably balanced distribution of classes when compared to the other tasks. The prompt learning setup comprised six combinations of a manual, mixed or soft template with a manual or soft verbaliser. The results are summarised in Table\ref{tab:prompt-combo}


\begin{table}[!htp]\centering
\caption{Table comparing different prompt learning setups on ICD9 Triage task.}\label{tab:prompt-combo}
\begin{tabular}{lrrr}\toprule
\textbf{PLM} & \textbf{Prompt combination} & \textbf{Balanced accuracy} \\\midrule
Fine-tuned & (manual, manual) &0.8765 \\
& (manual, soft) &0.8818 \\
& (mixed, manual) &0.8817 \\
& \textbf{(mixed, soft)}  & \textbf{0.8824} \\
& (soft, manual) &0.8860 \\
& (soft, soft) &0.8954 \\
\hline
Frozen & (manual, soft) &0.7524 \\
& (mixed, manual) &0.8474 \\
& \textbf{(mixed, soft)} & \textbf{0.8724} \\
& (soft, manual) &0.8591 \\
& (soft, soft) &0.8900 \\
\bottomrule
\end{tabular}
\end{table}
The performance across the different prompt combinations is very similar in the setting where the PLM is fine-tuned, however there is greater variance when the PLM is frozen. The frozen PLM setting is of most interest, and whilst the soft template and soft verbalizer combination performs the best overall, we opt to use the more interpretable combination of mixed template and soft verbalizer as our prompt learning benchmark going forward. The mixed template is a mixture of manual prompting and prefix tuning, whereby both discrete tokens known to the PLM and newly introduced, trainable continuous vectors of the same dimension as the PLM token embeddings are combined. 

\subsection{Prompt learning versus traditional fine-tuning}

\begin{figure}[!htb]
    \centering
    \includegraphics[width=0.99\textwidth]{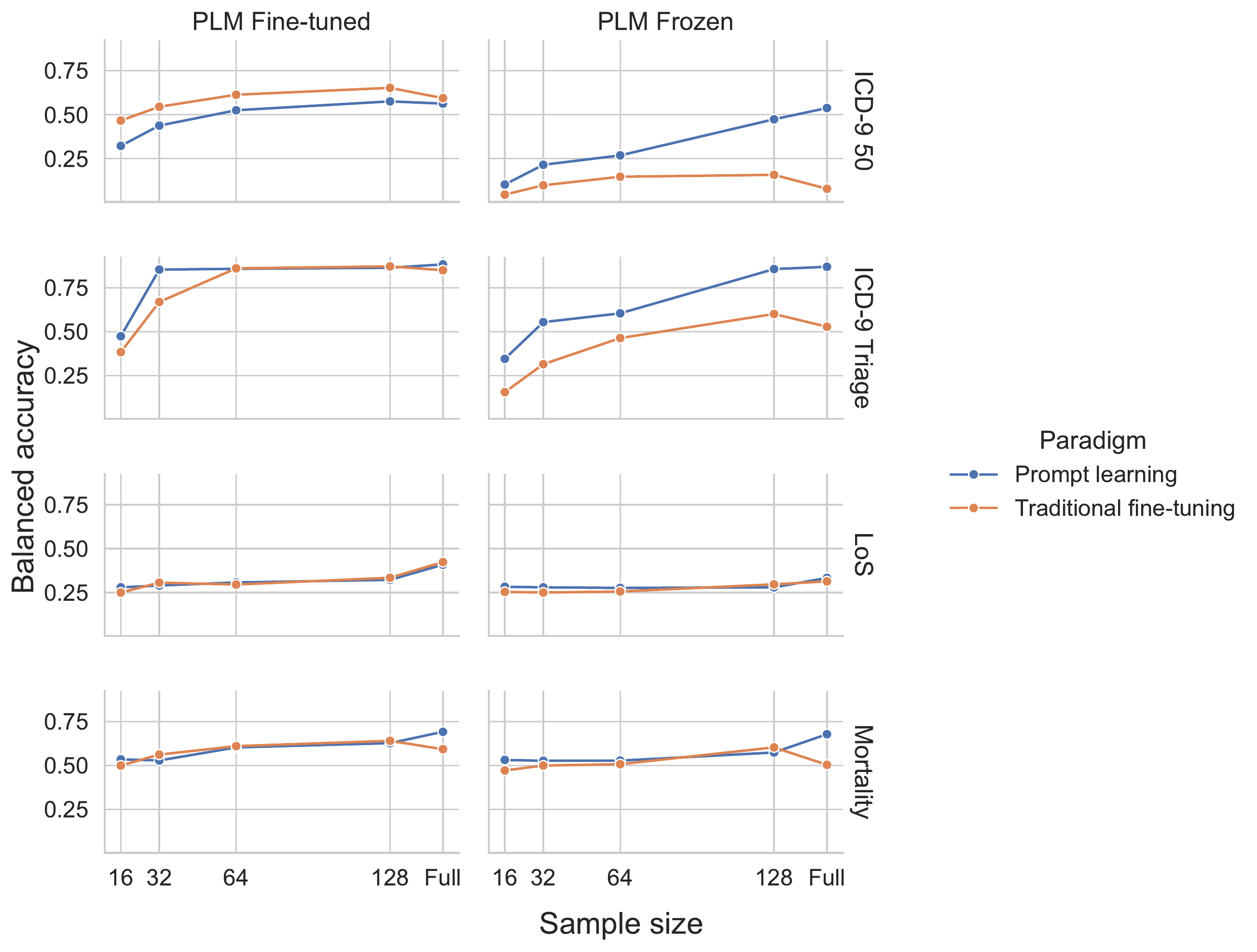}
    \caption{Balanced accuracy for prompt learning and traditional fine-tuning frameworks across the four clinical tasks. ``LoS'' refers to length of stay and ``Full'' refers to a full data set size which varies from task to task.}
    \label{fig:all-tasks-compare}
\end{figure}

Next is a comparison across the different clinical tasks outlined in the methods section between prompt learning and traditional fine-tuning. Each framework utilises the exact same PLM and we present evaluation results for both fine-tuning and freezing the entire PLM. In the case of the frozen PLM, only the parameters introduced by traditional fine-tuning or prompt learning are updated during training. We found that prompt learning can match or improve on traditional fine-tuning, with a much smaller gap in performance between the frozen and fine-tuned PLM setting across few-shot and full training setups, see Fig. \ref{fig:all-tasks-compare}.

\subsection{Hyperparameter search}
There are considerable variations in any neural networks performance with changes to hyperparameters, in particular learning rates and hidden layer dimensions. With comparing the performance of two neural network frameworks as we have, one must be careful to ensure the hyperparameters are optimized for each. Our initial experiments used sensible hyperparameters based on previous research using traditional fine-tuning and prompt learning, where prompt learning and traditional fine-tuning achieved similar performance when the PLM was fine-tuned, see Fig.\ref{fig:all-tasks-compare}. However, when freezing the PLMs, performance differences arose between the two frameworks, especially for few-shot settings in favor of prompt learning. We chose the ICD-9 Triage task as the optimal showcase task for further exploration due to its relatively stable performance. Moreover, with limited computational resources, it was impractical to run hyperparameter searches for all tasks and frameworks. The hyperparameter search space is provided below in Table \ref{tab:trad-hp-search}, with results of the subsequent optimized training runs for the ICD-9 Triage task presented in Table \ref{tab:optimized}. Further details of the hyperparameter search and results are presented in supplementary materials, see Appendix A. 


\begin{table}[!htp]\centering
\caption{Hyperparameter search space used for optmization}\label{tab:trad-hp-search}
\begin{tabular}{lrr}\toprule
\textbf{Parameter} &\textbf{Search space} \\
\midrule
classifier learning rate &log.uniform[$1 \times 10^{-5}, 3 \times 10^{-1}$] \\
batch size &[$4$] \\
gradient accumulation steps & range[$2, 10$] \\
dropout &range[$0.1, 0.5$] \\
optimizer &categorical[adamw, adafactor] \\
prompt learning rate &log.uniform[$1 \times 10^{-5}, 3 \times 10^{-1}$] \\
verbalizer learning rate &log.uniform[$1 \times 10^{-5}, 1 \times 10^{-1}$]\\
\bottomrule
\end{tabular}
\end{table}
%

\begin{table}[!htp]\centering
\caption{Hyperparameter optimized model comparison with frozen PLM for ICD9 triage.}\label{tab:optimized}
\begin{tabular}{lrrrrrr}\toprule
\textbf{Paradigm}  &\textbf{Balanced accuracy} &\textbf{F1 weighted} &\textbf{AUC} \\\midrule
Traditional fine-tuning  &0.8162 &0.8919 &0.9811 \\
Prompt learning &\textbf{0.8698} &\textbf{0.9246} &\textbf{0.9889} \\
\bottomrule
\end{tabular}
\end{table}

\subsection{Sensitivity analyses}
\begin{figure}[!htb]
    \centering
    \includegraphics[width=0.9\textwidth, scale=0.5]{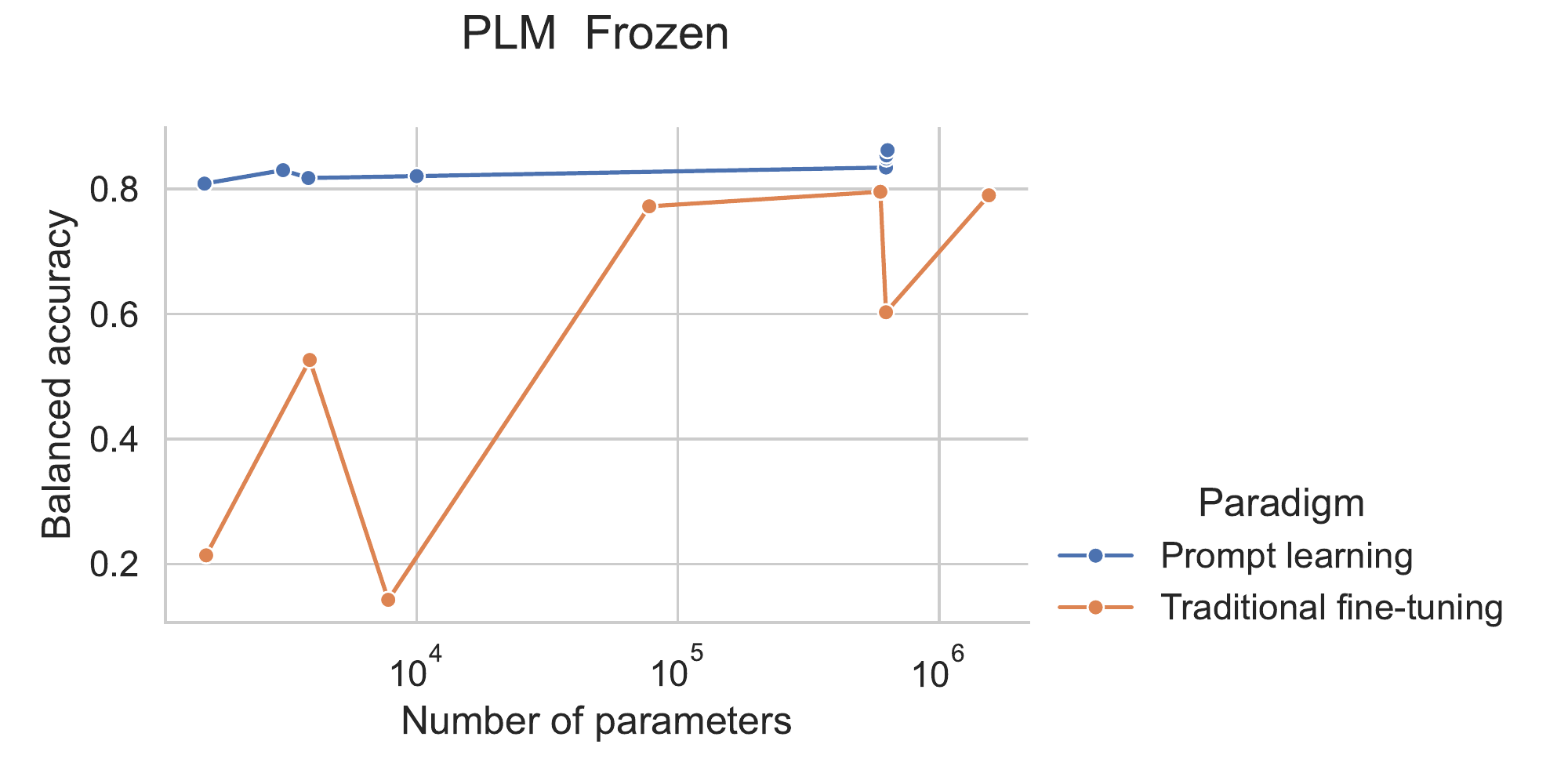}
    \caption{Balanced accuracy for prompt learning versus traditional fine-tuning across increasing number of trainable parameters with frozen PLM. For readability, logarithmic scale is used for $x$-axis.}
    \label{fig:sensitivity-analysis}
\end{figure}

Results suggested that on certain tasks prompt learning outperformed the traditional fine-tuning model when using a frozen PLM Fig.\ref{fig:all-tasks-compare}. We will focus on the triage task again, for which we optimized each of the frameworks. There is a risk that the performance drop for the traditional fine-tuning classification head is due to over or under fitting with its larger number of trainable parameters in the original setting. We manipulated the number of trainable parameters in each framework and compared the effects on performance, for results see Fig.\ref{fig:sensitivity-analysis}. Adjusting the number of trainable parameters for traditional fine-tuning involves adjusting the number of layers and hidden dimension size of the classification head, whilst adjusting number of trainable parameters for prompt learning requires just changing the number of soft template tokens and whether to include a soft verbalizer (manual templates and  verbalizers have no trainable parameters). Training used 128 samples per class as this approached peak performance without requiring a full training run.  Note that prompt learning with the \textit{fewest} trainable parameters (N params = 1,536) achieves comparable performance to the traditional fine-tuning model with $1000$ times the number of trainable parameters (N params = 1,552,007). 


The variability in prompt learning performance based on the template and verbalizer has been well established   \cite{pre-train-prompt-predict, prefix-tuning, Ding2021}. We opted to focus on the use of a mixed template format which is based around designing a common sense manual template for the task alongside soft and trainable tokens or embeddings. Moreover these soft tokens can be initialised from a known token of the PLM's vocabulary. To determine whether mixed templates benefit from a common sense or domain specific manual template, we compared performance of different templates including one with a mix of unrelated and random tokens. Results are shown in Table \ref{tab:prompt-compare} and we can see that having just one soft token or a set of random and unrelated manual tokens leads to a drop in performance. The <[soft]> token represents the trainable continuous vector or embedding of the mixed template that has been initialised from the PLMs vocabulary. Thus <[soft]>:"This" indicates a soft embedding initialised from the PLMs representation of the token "This". 

\begin{table}[!htp]\centering
\caption{Performance of the classification model on a test set for different mixed templates for the ICD9 triage task.}\label{tab:prompt-compare}
\begin{tabular}{p{7.5cm}r} 
\toprule
\textbf{Prompt text} &\textbf{Balanced accuracy} \\\midrule
{<[soft]>: "This"} {<[MASK]>} &0.8195 \\\midrule
{<[soft]>: "This"} patient {<[soft]>:"should go to"} {<[MASK]>}. &0.8539 \\\midrule
{<[soft]>: "This"} patient should {<[soft]>:"go to"} {<[MASK]>}. &0.8491 \\\midrule
{<[soft]>: "This"} patient should {<[soft]>:"go to this medical team based on symptoms of their illness"} {<[MASK]>}. &\textbf{0.8624} \\\midrule
random words here {<[soft]>:"random"} {<[MASK]>}. &0.8346 \\
\bottomrule
\end{tabular}
\end{table}

\section{Discussion}
The experiments presented here have attempted to directly compare the prompt learning paradigm with the traditional fine-tuning paradigm across a number of clinical tasks that frame classification as a clinical decision support task. The objective was to ascertain whether the literature describing promising performance for prompt learning in general domain text datasets can be leveraged on a more niche biomedical domain.  We present four clinical decision tasks of varying complexity, in both full training and few-shot setups. In the full training scenario, prompt learning can typically match the performance of traditional fine-tuning, and prompt learning outperforms traditional fine-tuning in the few-shot setting. Of particular interest was the performance of each model with frozen the PLM, where only parameters added to the PLM after pre-training are tuned for downstream classification tasks. This is where prompt learning appears to prove superior, out-performing traditional fine-tuning with considerably fewer trainable parameters, see Figure .\ref{fig:sensitivity-analysis}.  Moreover, the use of a mixed template appears to allow the intuitive common sense approach to domain derived prompts, whilst maintaining a trainable soft embedding that can reduce the difficulty in finding optimal manual prompts. We argue that mixed templates achieve similar performance to entirely soft templates, whilst retaining a level of transparency and interpretability. Understanding how models arrive at a decision is especially important in high-stake applications, such as medicine   \cite{taylor2021rationale, rajpurkar2022ai}. Future work should focus on the utility of interpretable prompts for helping clinicians understand a model's decision making.  

\subsection{Limitations}

\paragraph{Pre-training data leakage}
A notable limitation was the choice of PLM, which is arguably too well suited to the clinical tasks presented, with probable data leakage from initial pre-training and the subsequent downstream tasks. Although it must be stated that this would have benefited both paradigms, but there is the possibility that the reformulation of the downstream tasks as a masked language modelling style objective may allow easier "remembering" for prompt learning when compared to traditional fine-tuning. However, we include results for the ICD-9 Triage task using biomedical BERT (trained only on biomedical literature) and this yielded a similar pattern of results, see Appendix D. 

\paragraph{Task performance variance}
We presented four clinical tasks derived from MIMIC-III notes data, and whilst we achieved results in line with previous research, the relative performance on the length of stay and mortality prediction tasks were quite poor regardless of the framework. This limits the interpretability of framework differences in performance, and whether one is more suitable to some tasks than others. Similarly we did find that using hyperparameter search for the ICD-9 Triage task improved the frozen PLM performance of the traditional fine-tuning approach by a reasonable margin and a more extensive hyperparameter search may shift this further. However, this was also true for the prompt learning approach, but these models appeared far more robust to changes in hyperparameters. Future work would benefit from exploring this more extensively, given adequate computing resource.

\subsection{Conclusion}
The key finding was that prompt learning outperforms the traditional fine-tuning approach when PLMs are frozen during training on the downstream task. Most striking is the relatively few trainable parameters required for prompt learning to converge and match or even outperform traditional fine-tuning. This is in line with previous prompt learning research and may offer a useful framework for building clinical support tools in low compute resource settings, as well as enabling a faster, flexible, modular training pipeline for new downstream tasks and novel data. The ability to utilise a single, frozen PLM and share or reuse these embeddings across a number of task specific modules, each with their own trainable prompt is very desirable for specialised domains. Whilst using smaller PLMs and prompts may not achieve the state of the art performance on certain tasks, it can approach similar levels of performance with a fraction of the model size and training time. In the field of clinical support tools, a computationally efficient and interpretable model with good enough performance that can run on a CPU is arguably more desirable than a trillion parameter model that requires high-performance computing clusters with arrays of GPUs. The prompt learning framework is an evolving paradigm with variants being introduced regularly, thus we cannot claim to have fully covered prompt learning in this work. We have opted to use the most readily available, and arguably resource efficient prompt approach to achieve our results. This work can act as a basis for further clinical prompt learning work, and may encourage the use of relatively small domain specific PLMs rather than relying on the giant PLMs produced by commercial enterprises. We suggest that it is more efficient to train a small BERT model on a specialised domain and applying prompt learning, than attempting to apply prompt learning directly to models such as GPT-3 which often lack the domain knowledge required.   

\section*{Acknowledgement}

NT is supported by the EPSRC Center for Doctoral Training in Health Data Science (EP/S02428X/1). AK, ANH, YZ and DWJ were supported in part by the NIHR AI Award for Health and Social Care (NIHR-AI-AWARD0-2183); AK and ANH declare a research grant from GlaxoSmithKline. DWJ is supported by the NIHR Oxford Health Biomedical Research Centre (grant BRC-1215-20005).  The views expressed are those of the authors and not necessarily those of the UK National Health Service, the NIHR, the UK Department of Health, or the University of Oxford.

\bibliographystyle{unsrtnat}
\bibliography{references}

\appendix
\section*{Appendix} 
\counterwithin{figure}{section}
\counterwithin{table}{section}
\section{Training details}

We implement our experiments using a combination of the OpenPrompt framework  \cite{Ding2021} and the Pytorch packages. For prompt learning, we use Adafactor \cite{adafactor} optimizer for soft and mixed templates, and AdamW  \cite{adamw} optimizer for language models and soft verbalizers. For traditional fine-tuning, we use AdamW optimizer for MLP heads and language models. We train the model on a Nvidia RTX 1080 Ti GPU, with a batch size of 4 due to the memory limitation. To overcome this, we use gradient accumulation for 10 steps during training. Further details of training and hyperparameters can be in the complimentary code repository.

Table \ref{tab:optim-hparams} shows the derived optimal hyperparameters for each training paradigm based on the hyperparameter random search. The search consisted of 100 training runs using randomly generated hyperparameters from the search space shown in Table \ref{tab:trad-hp-search}. Due to relatively limited computational resource, this was only performed for the ICD-9 Triage task and a sub-sample of the training data was used, similar to that of our few-shot experiments with 128 samples per class.

\begin{table}[!htp]\centering
\caption{Optimized hyperparameters for each training paradigm}\label{tab:optim-hparams}
\begin{tabular}{lrrr}\toprule
\textbf{hp} &\textbf{Traditional fine-tuning} &\textbf{Prompt learning} \\\midrule
learning rate &0.0048 &0.0121 \\
batch size &4 &4 \\
gradient accumulation steps &4 &3 \\
dropout &0.382 &0.1536 \\
optimizer &adamw &adafactor \\
verbalizer learning rate &n/a &0.007 \\
\bottomrule
\end{tabular}
\end{table}

\section{Dataset details}

\paragraph{Mortality and Length of Stay}
For all clinical tasks a combination of available clinical notes pertaining to the outcome of interest were used, including admission and discharge summaries. Each task dataset was created separately and a 70-10-20 split of training-validation-test sets was used. We followed the data engineering steps outlined in the clinical outcomes paper   \cite{van-aken-etal-2021-clinical}. 

\paragraph{ICD-9 50 and ICD-9 Triage}
The ICD-9 50 task was simply all clinical notes data corresponding to the top 50 most frequently occuring ICD-9 diagnosis codes. The production of the ICD-9 Triage task was derived from taking the top 20 ICD-9 diagnosis codes. From this subsample, a clinician derived suitable groups representing the destination team on discharge from ICU: Cardiology, Obstetrics, Respiratory Medicine, Neurology, Gastroenterology, Acute or Internal Medicine, and Oncology. 

\begin{figure}[!htb]
    \centering
    \includegraphics[width=0.9\textwidth]{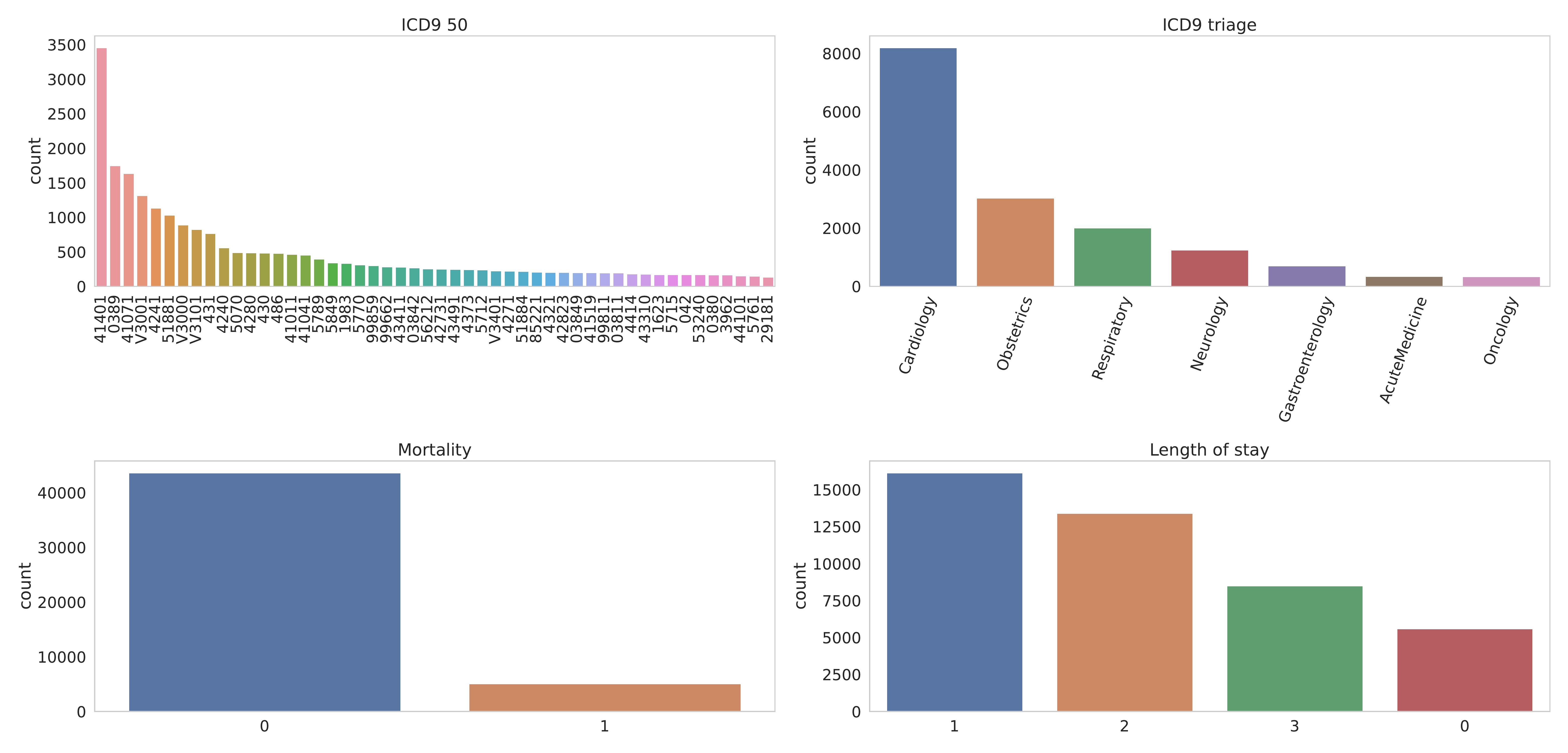}
    \caption{Distribution of classes for each clinical task}
    \label{fig:all-dist}
\end{figure}

See Fig.\ref{fig:all-dist} showing class distributions for each of the clinical tasks presented in this paper.

\section{Prompt examples}

Examples of different prompt methods are shown. For each task we show one manual prompt template and one mixed template. The <[soft]> token represents the trainable continuous vector or embedding of the mixed template that has been initialised from the PLMs vocabulary. Thus <[soft]>:"This" indicates a soft embedding initialised from the PLMs representation of the token "This". 

\paragraph{ICD-9 diagnosis code triage}	

\begin{itemize}
\item <clinical note> Best department is {<[MASK]>}.	
\item <clinical note> {<[soft]>: "This"} patient should {<[soft]>:"go to this medical team based on symptoms of their illness"} {<[MASK]>}.
\end{itemize}

\paragraph{Mortality prediction}	

\begin{itemize}
\item <clinical note> Patient is on the path to {<[MASK]>}.
\item \begin{minipage}[t]{0.95\linewidth}<clinical note> {<[soft]>}: "This" patient {<[soft]>}:"on path to" {<[MASK]>}.\end{minipage}
\end{itemize}

\paragraph{ICD-9 diagnosis code classification - top 50}	

\begin{itemize}
\item <clinical note> Patient has diagnosis {<[MASK]>}	
\item \begin{minipage}[t]{0.9\linewidth}<clinical note>  {<[soft]>: "This"} patient <[soft]>:"has diagnosis" <[MASK]>.\end{minipage}
\end{itemize}

\paragraph{Length of stay prediction}	

\begin{itemize}
\item <clinical note> The patient will be at hospital with a {<[MASK]>} length.	
\item <clinical note> {<[soft]>: "This"} patient {<[soft]>:"will be in hospital for a "} {<[MASK]>} length.
\end{itemize}

\section{Prompt learning versus Traditional fine-tuning with PubMed BERT}
The PLM used for all presented results in the main body of the paper was the Bio-ClinicalBERT  \cite{alsentzer-etal-2019-publicly}, which we have observed was trained using Mimic-III notes. Whilst this was arguably advantageous for both traditional fine-tuning and prompt learning, it may have overly favoured prompt learning due to the reformulation of the classification task as a Masked Language Modelling (MLM) objective. Therefore we present results of another biomedical BERT model from Microsoft, the PubMedBERT, which was pre-trained from scratch using abstracts from PubMed  \cite{pubmedbert} in Table \ref{tab:microsoft-bert-results}. It can be seen that prompt learning still outperforms traditional fine-tuning by a large margin on the ICD-9 Triage task, in line with our other results.

\begin{table}[!htp]\centering
\caption{Balanced accuracy results for prompt learning and traditional fine-tuning using Microsoft's PubMedBert}\label{tab:microsoft-bert-results}
\begin{tabular}{lrrr}\toprule
&\multicolumn{2}{c}{\textbf{Balanced Accuracy}} \\\cmidrule{2-3}
\textbf{Sample size} &\textbf{Traditional fine-tuning} &\textbf{Prompt learning} \\\midrule
16 &0.1554 &0.2249 \\
32 &0.1521 &0.3749 \\
64 &0.4048 &0.4621 \\
128 &0.5621 &0.7814 \\
\bottomrule
\end{tabular}
\end{table}

\end{document}